\pdfoutput=1

\documentclass[11pt]{article}

\usepackage{acl}

\usepackage{times}
\usepackage{latexsym}

\usepackage[T1]{fontenc}

\usepackage[utf8]{inputenc}

\usepackage{microtype}
\usepackage{comment}
\usepackage{xcolor}
\usepackage{todonotes}
\usepackage{booktabs}
\usepackage{nicematrix}

\usepackage{multirow}
\usepackage{rotating}
\usepackage{array, graphicx}
\usepackage{cleveref}

\title{PizzaCommonSense: Learning to Model Commonsense Reasoning about Intermediate Steps in Cooking Recipes}

\author{A\"issatou Diallo$^1$\thanks{ Corresponding author: \texttt{a.diallo@ucl.ac.uk}\\}, Antonis Bikakis$^2$, Luke Dickens$^2$, Anthony Hunter$^1$, Rob Miller$^2$ \\
$^1$Department of Computer Science \\
$^2$Department of Information Studies     \\
University College London, United Kingdom \\}

\begin{document}
\maketitle
\begin{abstract}

Understanding procedural texts, such as cooking recipes, is essential for enabling machines to follow instructions and reason about tasks, a key aspect of intelligent reasoning. 
In cooking, these instructions can be interpreted as a series of modifications to a food preparation.
For a model to effectively reason about cooking recipes, it must accurately discern and understand the inputs and outputs of intermediate steps within the recipe.
We present a new corpus of cooking recipes enriched with descriptions of intermediate steps that describe the input and output for each step. PizzaCommonsense serves as a benchmark for the reasoning capabilities of LLMs because it demands rigorous explicit input-output descriptions to demonstrate the acquisition of implicit commonsense knowledge, which is unlikely to be
easily memorized. GPT-4 achieves only 26\% human-evaluated preference for generations, leaving room for future improvements.
\end{abstract}

\section{Introduction}

Procedural text is a type of writing that provides instructions on how to perform a task using resources to achieve a final goal. Common real-world examples include scientific articles, DIY instruction books, or cooking recipes \cite{tang2020understanding,gupta2019effective,gupta2019tracking}. In the latter case, the procedural text instructs an agent on how to prepare a dish. To understand and follow a recipe, one must be able to reason about the steps involved and the effects of each step on the ingredients. This requires common sense knowledge about cooking, such as knowing how different cooking techniques and food properties affect the final product.
\begin{figure}[!t]
    \centering
    \includegraphics[width=7.5cm]{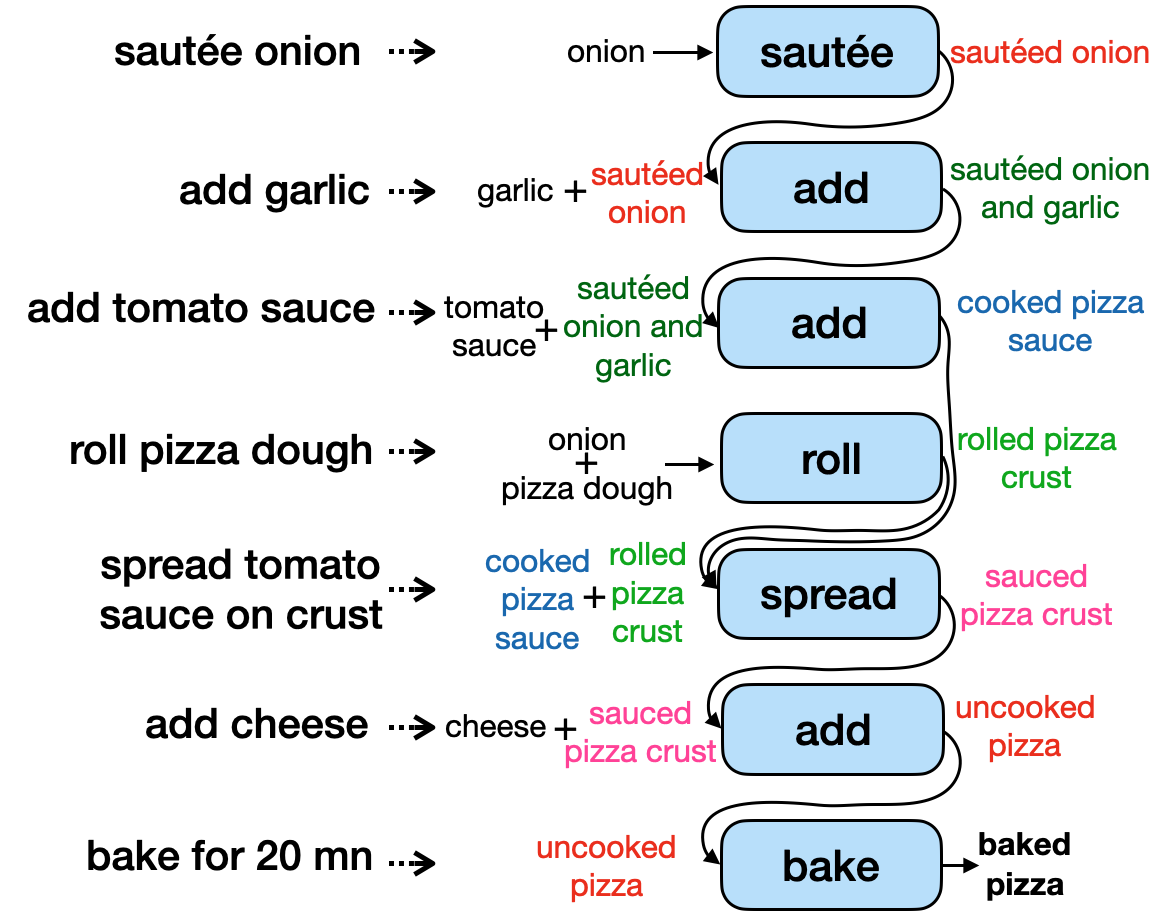}
    \caption{A graphical depiction of the PizzaCommonsense underlying motivation. Models are required to learn knowledge about the input and output of each intermediate step and predict the correct sequencing of these comestibles given the corresponding instructions and cooking actions.}
    \label{fig:recipe example}
\end{figure}
Humans can easily imagine the effects of each step in a recipe as they read it, even if they have never prepared the dish before. 
They can also use commonsense to reason about the recipe, the purpose of the action in the instruction, as well as the input and output of the cooking step, determining what comestibles are necessary for performing a specific step, predicting and understanding the effects of performing a cooking action, giving explanations about the undertaken actions, identifying alternative orderings, and adapting to the cooking conditions.

With the advent of increasingly capable artificial tools, such as Large Language Models (LLMs) comes the need to investigate their commonsense reasoning abilities in following procedural text such as cooking recipes. Inspired by the recent line of work of prompting LLMs \cite{kojima2022large,wei2022chain,wang2022self} to generate a reasoning chain along with the answer, with the goal of mimicking the human reasoning process, we argue the need to evaluate the correctness of the intermediate reasoning steps.

For this reason, we propose PizzaCommonSense, a dataset for commonsense reasoning about intermediate and implicit steps for cooking recipes. A visual representation of the purpose of the data set is given in \Cref{fig:recipe example}. This dataset contains pizza recipes that are parsed into atomic cooking steps such that each step contains only one cooking action. The recipes are organized in a tabular format with four different columns, the cooking instruction, the input preparation, the relevant cooking action and the output preparation. The task is at the interface between question answering (QA) and natural language inference. Given the set of instructions of a recipe, models are required to predict a description for the input and output preparations for each cooking step.

This is particularly challenging because models are required to reason and predict accurate descriptions of the intermediate steps. The intermediate steps of a cooking recipe are the steps that are performed after the initial preparation and before the final plating and presentation of the dish. These steps are typically where the main cooking and transformation of the ingredients take place. 


Concretely, the task involves individuating the (i) explicit input comestibles (e.g. \textit{"sauté the \underline{onion} in the skillet")} and (ii) output comestibles (e.g. \textit{"put the \underline{tomato sauce} in a bowl")}. Natural language makes often use of omissions and anaphoras which a good model should be able to resolve by identifying (iii) implicit input comestibles (e.g. \textit{"season the sauce to taste"} with the implicit input being "\underline{seasoning}" while "\underline{sauce}" is explicit) and implicit output comestibles (e.g. \textit{"mix the flour, water, salt and yeast"} where the implicit output is "\underline{dough}"). 

We propose baseline methods to solve the task: T5 \cite{raffel2020exploring} with fine tuning; Flan-T5 \cite{chung2022scaling} with prompting and fine tuning, GPT-3.5 with demonstrations, fine-tuned GPT-3.5 and GPT-4 with Chain-of-thought (CoT) prompts.

The contributions of this work are threefold: (1) we propose a new task for procedural text comprehension, namely predicting the input and output of a given action while giving self-contained descriptions of the transformed resources for each step of the procedural text; (2) we construct an annotated dataset to facilitate the studies of commonsense reasoning for procedural texts with the dataset being made publicly available;
and 
(3) we benchmark the performance of state-of-the-art generative language models on our dataset and demonstrate the difficulty of the task.

\section{Motivation}

Commonsense reasoning is central to human intelligence. It is essential for humans to operate in the real-world. In AI, we lack a full understanding of commonsense reasoning or the means to simulate it. By creating datasets that reflect domains suitable for investigating commonsense reasoning, we can explore modeling techniques. 

Cooking, for instance, is a domain where commonsense reasoning is vital. Consider developing a robot to cook; it needs to deeply understand recipe instructions, which involves interpreting ambiguous language, handling missing information, and resolving references common in procedural texts. To teach a robot to cook effectively, it must understand recipes deeply, including ingredients, final products, and intermediate states. Recognizing these intermediates and their properties is crucial for appropriate actions—for instance, suppose a robot is using a pizza recipe and it reads the instruction to mix 20oz of flour with 10oz of water, it needs to use commonsense reasoning that the result is dough (often this is not explicitly stated), and then inferring that it is dough, the robot can decide that it can move this intermediate comestible around the kitchen by hand. Later in the recipe, suppose the robot reads that it needs to put 10oz of tomatoes into the blender. In this case, the robot needs to use commonsense reasoning to determine that the intermediate comestible will be tomato puree, and that it cannot move it in its hands, but rather requires a receptacle. 

Achieving human-level cooking requires an agent to understand nuances of each step and anticipate outcomes, translating ambiguous recipe language into precise instructions. 
 
Our dataset uniquely provides annotated recipe instructions, labeling intermediate input and output comestibles. By focusing on the implicit transformations between these elements, we aim to bridge a critical gap in research and develop models that can truly understand and execute complex cooking tasks. This can then enhance the  capabilities of autonomous agents in various roles, including robotic cooking and virtual assistants.


\section{PizzaCommonsense}
\paragraph{Why pizza recipes?} The choice of focusing on pizza recipes stemmed from their inherent compositional nature within a controlled setting. In fact, pizza preparations typically involves a limited set of steps: creating the base, spreading a sauce, adding toppings, and baking/chilling/setting. This breakdown makes pizza an ideal candidate for studying the compositional aspects of recipe instructions. Each stage (base, sauce, toppings) acts as a distinct "building block", allowing for easier segmentation and analysis of the construction and reasoning process.

Recipes often provide high-level instructions that necessitate deconstruction into precise, low-level actions. This process is significantly influenced by the characteristics of intermediates. While our dataset focuses on pizza recipes, we believe that methods developed for commonsense reasoning with intermediates can be generalized to other cooking recipes, and then to other situations involving actions and intermediates.


\paragraph{Data sources}
We build our dataset on Recipe1M by \citet {salvador2017learning} which consists of one million structured cooking recipes with 13M associated images. We sample a set of 1087 recipes that contain the word "pizza" in the title and heuristically remove exact duplicate recipes. It is worth mentioning that despite the choice of selecting the recipes explicitly mentioning the word pizza, not all recipes have as a final product a pizza. 
Additionally, we use a cooking actions glossary from the dataset Now You're Cooking!\cite{kiddon2016globally,bosselut_simulating_2018}. This glossary lists the most common cooking actions that involve a change of state of the food items.
\paragraph{Pre-processing}
Our goal is to create atomic instructions, where each sentence depicts a single cooking action. We achieve this by parsing cooking steps and splitting only coordinate sentences. For instance, \textit{"Heat and stir to mix evenly"} becomes two separate atomic instructions: "Heat" and "Stir to mix heavenly." Here, \underline{"heat"} is the root verb identified by the dependency label \underline{ROOT}. \underline{"Stir"} connects to the root via the coordinating conjunction "and,". Conversely, "mix" is preceded by the particle "to" and acts as a clause modifier, modifying the verb \underline{"stir"}. In another example, the sentence \textit{"after you heat, stir to mix evenly"} remains intact. "After you heat" functions as a prepositional phrase, not a coordinate clause. In short, we rely on the combined analysis of three elements: (i) analyzing the syntactic dependency \textit{conj}; (ii) examining the dependency tag of the verb (and subject) in the second part of the conjunction; (iii) checking if the hypothetical verb in the second part is a cooking action listed in the cooking action glossary.
This step aims to promote reasoning about an action's purpose and the necessary intermediate transformations to prepare a dish. Hence, input-output pairs are meaningful only for actions that contribute to a state change in the comestibles.

\paragraph{Annotation collection}

We ran the annotation process on Amazon Mechanical Turk (AMT). 
We choose to frame the data collection process in a tabular format which is well adapted to the task for the clarity and conciseness and makes it easier to understand the relationships between the information in the columns and the different rows. Organizing the information using this format allows easy identification and analysis of the relationships between input and output, as well as those between different instructions. 

The columns of interest are: \textit{Instruction, Input, Action}, and \textit{Output}. We ask the crowd-workers to provide clear and understandable descriptions of the intermediates given the action and the instructions. We enumerate some constraints: (i) the "input" cell represents the state of the food preparation before the cooking action is performed and the "output" cell represents the state of the food preparation after the cooking action is performed; (ii) for steps that do not refer to any comestibles (e.g. \textit{preheat the oven to 450F}), "NA" is used for both the input and the output. Comestibles in a set are separated by a semicolon; 
(iii) verbs of motion such as \textit{move, place, transfer} have identical input and output.

The first row of each table is pre-filled, if the cooking action is a verb whose object is not a comestible with "NA". 
In addition, the crowd-workers performing the HIT were asked to not insert numerical values and if the output of a previous step were to become the input of a following instruction, to keep the description the same. The upper part of \Cref{fig:pipeline} summarizes the annotation collection. Details of the annotation interface are shown in the Appendix.

\paragraph{Key statistics}
\Cref{tab:key_statistics} shows the key statistics of PizzaCommonSense. 
The dataset contains 13141 data instances of instruction, input, cooking action and output among 1087 annotated recipes. An annotated example is shown in \Cref{fig:pipeline}. The average completion time per recipe is 5 minutes. We do not collect personal information about the crowd-workers. Crowd-workers were from Canada and United States. We manually checked the annotated recipes to ensure the quality of the collected data.

\paragraph{Distribution-based split} 
To preserve ingredient distribution across training, validation, and test sets, we implement distribution-based data splits via clustering. Initially, we extract recipe ingredients, removing specific terms like brands and quantities before vectorization. The clustering algorithm then iteratively merges recipes into clusters based on a distance metric. Recipes are assigned from each cluster to corresponding splits to ensure accurate representation and evaluation.

\section{Task}
We extend procedural text comprehension by additionally identifying each instruction step's intermediate input and output. This involves predicting both the input and output for each instruction, including implicit and explicit ingredients, action outcomes, and each step's resulting condition. We posit that this capability will generate faithful recipe variations and enhance reasoning about the recipe’s structure and components.
 

\begin{figure*}[!t]
\centering
\includegraphics[width=\textwidth]{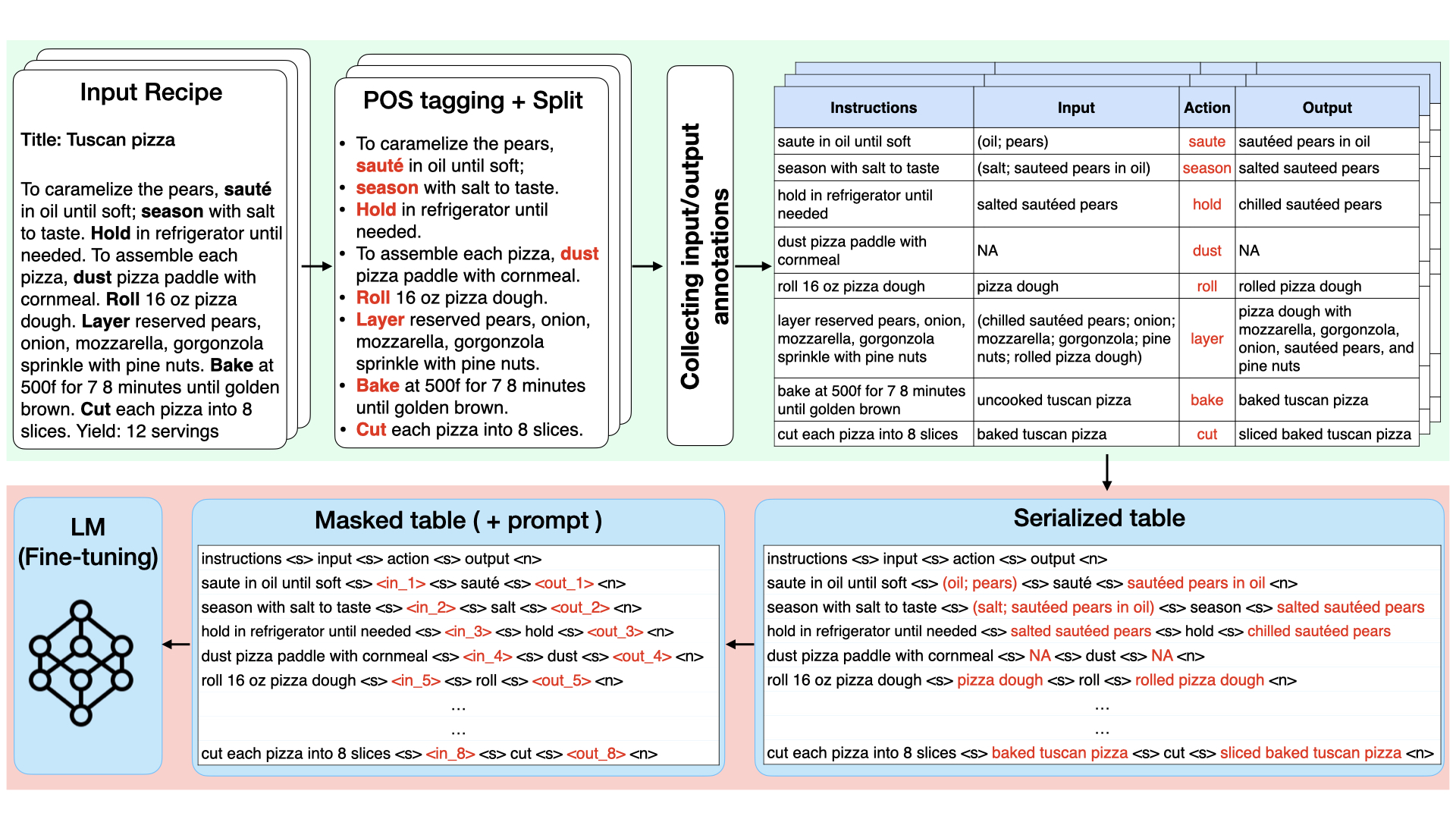}
\caption{Our proposed pipeline to obtain PizzaCommonSense. Given a recipe among the selected ones from Recipe1M, we first apply POS tagging to identify the cooking actions and split the sentences such that each sentence contains only one main cooking action. The instructions and the identified cooking action are formatted into a table which becomes the HIT.The green box illustrates the annotation process, and the red box represents the training/inference phase.}
\label{fig:pipeline}

\end{figure*}

\paragraph{Problem Formalization} We have a tabular representation of a recipe $R$ with $n=4$ columns (\textit{Instructions, Input, Action, Output}), 
and $m$ rows, one for each instruction of the recipe. 
Additionally, (i) the output cell of the row $t-1$ can be the content of the input cell of the row $t$ and (ii) the input and output cells of a row $t$ can be the same, if the cooking action is not transformative. 
The goal is to predict the content of the columns \textit{Input} and \textit{Output}, given the content of \textit{Instruction} and \textit{Action}. 

\paragraph{Input Representation} 
To use text-to-text generation baseline, we convert tables into a text sequence. We define a function \texttt{serialize(R)} that takes as input the tabular representation of the recipe $R$ and outputs a textual representation of the input. To handle the missing values in the table, the function \texttt{mask} fills the missing values with a mask token according to the architecture at hand. For example, for the baseline T5, we rely on the predefined sentinel tokens for the original span denoising objective. Otherwise, we use standard masking tokens \texttt{<in>} and \texttt{<out>} to indicate the missing values. The \texttt{<s>} token is used to separate the cells of each row, \texttt{<n>} is used to separate the rows in the serialized table. 
The content of the \textit{input} and \textit{output} cells can be a set of comestibles or ingredients. For example, the instruction explaining how to assemble a dish is likely to have multiple  food items that need to be combined. In order to ensure clarity and readability, the ingredients are listed between brackets and separated by semicolons. An example of this is given in Fig.\ref{fig:pipeline}.

\subsection{Models} We evaluate baselines with four models: T5-base \cite{raffel2020exploring} with 220M parameters, Flan-T5-base \cite{chung2022scaling} with 250M parameters, GPT-3.5 \cite{openai2021chatgpt} and GPT-4 \cite{achiam2023gpt}. 

\begin{table*}[!t]
\centering
\caption{Quantitative evaluation for T5-based models under distribution based splits and random splits. EMA is exact matching accuracy, B is Bleu, R is Rouge$_L$, M is Meteor and BS stands for BertScore. Higher is better.}
\label{tab:t5_results}
\scalebox{0.9}{
\begin{NiceTabular}{@{}lclllll|clllll@{}}
\toprule
                                                & \multicolumn{6}{c|}{\textbf{Random}}                                                                                                                        & \multicolumn{6}{c}{\textbf{Clustering}}                                                                                                                    \\ \midrule
                                                & \multicolumn{2}{c}{\textbf{Input}}                        & \multicolumn{4}{c|}{\textbf{Output}}                                                            & \multicolumn{2}{c}{\textbf{Input}}                        & \multicolumn{4}{c}{\textbf{Output}}                                                            \\ \cmidrule(l){2-13} 
                                                & EMA                               & \multicolumn{1}{c}{R} & \multicolumn{1}{c}{B} & \multicolumn{1}{c}{R} & \multicolumn{1}{c}{M} & \multicolumn{1}{c|}{BS} & EMA                               & \multicolumn{1}{c}{R} & \multicolumn{1}{c}{B} & \multicolumn{1}{c}{R} & \multicolumn{1}{c}{M} & \multicolumn{1}{c}{BS} \\ \midrule
\multicolumn{1}{r}{\textbf{Fine-tuned T5}}      & \multicolumn{1}{l}{16.2}          & \textbf{58.5}         & \textbf{24.67}        & 54.5                  & 45.8                  & 78.9                    & \multicolumn{1}{l}{12.9}          & 48.3                  & 15.1                  & 44.5                  & 36.7                  & 74.5                   \\ \midrule
\multicolumn{1}{r}{\textbf{Fine-tuned Flan T5}} & \multicolumn{1}{l}{\textbf{16.7}} & 53.5                  & 21.3                  & \textbf{54.6}         & \textbf{45.9}         & \textbf{86.9}           & \multicolumn{1}{l}{\textbf{13.8}} & \textbf{53.5}         & \textbf{17.7}         & \textbf{50.4}         & \textbf{41.5}         & \textbf{83.3}          \\ \bottomrule
\end{NiceTabular}
}
\end{table*}

\paragraph{Fine Tuning T5-based models}
We use the pretrained objective of T5 and Flan-T5 and we build a baseline through fine tuning the pretrained T5-base model in a sequence to sequence fashion. For doing this, we use as input the serialized table with the \textit{input} and \textit{output} masked out and the output is the content of corresponding input and output. 


\paragraph{GPT-3.5+demo} 
We test in-context learning setting for predicting the input/output pairs. To do this, for each test sample (\texttt{serialize(R)}, \texttt{serialize(mask(R))}) from the annotated dataset, we sample a (for 1-shot setting) serialized recipe table from the training set which is used as demonstration for GPT-3.5 to perform the task. 

\paragraph{GPT-3.5+FT} Fine-tuning tailors the language model's capabilities to specific tasks and domains. 
We follow the same structure used for fine-tuning T5 where the \textit{input} and \textit{output} are masked with special tokens \texttt{<in>} and \texttt{<out>}. 
\paragraph{GPT-4+ CoT} Chain-of-thought prompting (CoT) is a technique that consists of appending \textit{"Let's think step by step"} at the end of the instruction which improves the performance by making the reasoning steps explicit. 

\subsection{Preliminary study} 
We first conduct a preliminary study to inspect the performance of the chosen baselines without fine-tuning. After serializing the table, we feed each table in our annotated test set to the baseline models. From Table \cref{tab:preliminary} we can observe that all models (except for GPT-4) perform poorly out-of-the-box.
\begin{table}[h]
\centering
\caption{0-shot results without finetuning. EMA stands for exact matching accuracy, R stands for Rouge$_L$ and BS stands for BertScore.} 
\label{tab:preliminary}
\scalebox{1}{
\begin{tabular}{@{}lllll@{}}
\toprule
        & \multicolumn{2}{c}{\textbf{Input}}              & \multicolumn{2}{c}{\textbf{Output}}            \\ \cmidrule(l){2-5} 
        & \multicolumn{1}{c}{EMA} & \multicolumn{1}{c}{R} & \multicolumn{1}{c}{R} & \multicolumn{1}{c}{BS} \\ \midrule
T5      & 0.2                     & 0.71                  & 0.84                  & 59.9                   \\
Flan-T5 & 0.0                     & 0.0                   & 0.64                  & 26.2                   \\
GPT-3.5 & 7.6                     & 26.5                  & 27.8                  & 87.1                   \\
GPT-4   & \textbf{18.6}           & \textbf{41.4}         & \textbf{52.3}         & \textbf{89.5}          \\ \bottomrule
\end{tabular}
}

\end{table}
In particular, T5 models fail to predict appropriate text. Conversely, GPT-3.5 incorrectly formatted predictions in 72\% of cases, either paraphrasing or rewriting instructions, predicting the title or final recipe object, or placing correct placeholders in a different format. An example is shown in the supplementary material. 
These are excluded when computing preliminary results.

\section{Experimental setup}
\paragraph{Parameters}

We use T5-base and Flan T5-base models with a batch size of 16 and a learning rate of $1e^{-5}$, using Adam optimizer \cite{kingma_auto-encoding_2013}. We fine-tune for 30 epochs and save the checkpoint with the best performance on the validation set. Both T5 based models are pretrained in English. Both T5 and Flan T5 employ around an hour on 4 GTX GPUs. All values are the averaged across 3 runs.
For the LLM-based methods, we use \texttt{gpt-3.5-turbo-1106} and \texttt{gpt-4-turbo} models with temperature set at 0. The cost of finetuning the GPT-3.5 was 15 USD.
\paragraph{Automatic evaluation}
Our evaluation uses the same protocol as \cite{lin2020commongen}. Specifically, we use 
Bleu \cite{papineni2002bleu}, Rouge$_L$ \cite{lin2004rouge}, Meteor \cite{banerjee2005meteor}. To assess the validity of the generated outputs, we include BERTScore \cite{zhang2019bertscore}, a content-oriented and semantic metric. Due to the different nature of the intermediate inputs and the intermediate outputs we differentiate the evaluation of the two predicted elements. We use exact matching accuracy (EMA) \cite{keysers2019measuring,kim2020cogs}, which computes the percentage of instances where two strings match exactly, and Rouge$_L$ for the quantitative evaluation of the input. The intermediate outputs are rarely described as sets of comestibles, and are much closer to natural language descriptions, so evaluated with Rouge$_L$, Bleu, Meteor and BertScore. 

\paragraph{Human evaluation}
We supplement this protocol with a fine-grained human evaluation. We present 50 generations to a set of 25 recruited evaluators different from the annotators. We evaluate the generated outputs based on four criteria: (1) \textbf{Completeness}: true if all the relevant comestibles are present in the generated input; (2) \textbf{Validity}: true if and only if the generated text follows all the rules ("only comestibles", "descriptive predictions"); (3) \textbf{Consistency}: true if the input-output pair makes logical sense; (4) \textbf{Win/Tie/Lose}: the generated outputs are compared to the gold reference to determine whether they are preferred (Win), equivalent (Tie), or less preferred (Lose).


\section{Results}
We analyze the performance of the baseline models in predicting the correct input and output for the intermediate steps of a cooking recipe. To perform well, the models should predict (i) only comestibles in the input/output pairs; (ii) all the comestibles necessary to perform the cooking action at the given time step, implicitly and explicitly stated in the cooking instruction. In particular, it should not fail to include the output of the previous step if the next cooking instruction is a transformation of the comestible; (iii) the comestibles implicitly mentioned could also be inferred by the cooking action, i.e. the input associated with \textit{Salt the sauce} should be \textit{(salt; sauce)} and finally (iv) specific descriptions of the output comestibles, i.e. although not strictly wrong, describing a sauce as \textit{a mixture of tomato paste, oil, water and seasonings} is undesirable.

\begin{table*}[!t]
\centering
\caption{Fine-grained human evaluation. Overall consistency is marked on a binary scale– the input-output relation correct. Completeness penalizes for missing elements and validity measures if the generated sequence follows the given rules. For Win/Lose/Tie, annotators compared the generations against the gold references. Last row is human performance for comparison.}
\label{tab:human_eval}
\scalebox{1}{
\begin{tabular}{@{}lcccccc@{}}
\toprule
                      & Completeness  & Validity      & Consistency  & Win ($\uparrow$) & Tie  & Lose ($\downarrow$) \\ \midrule
\textbf{Flan T5 + FT} & 0.52          & 0.42          & 0.82         & 0.04             & 0.04 & 0.92                \\
\textbf{GPT-3.5 + FT} & \textbf{0.72} & \textbf{0.72} & \textbf{1.0} & 0.24             & 0.02 & 0.74                \\
\textbf{GPT-4 + CoT}  & 0.68          & 0.68          & 0.95          & 0.26             & 0.06 & 0.68                \\ \midrule \midrule
\textbf{Human}       & 0.95 &   0.98       &1.0     &   0.3   & 0.65          & 0.05                 \\ \bottomrule
\end{tabular}}
\end{table*}

\begin{table}[ht]
\centering
\caption{Quantitative evaluation for GPT-3.5+demo, GPT-3.5+FT and GPT-4+CoT. EMA is exact matching accuracy, B is Bleu, R is Rouge$_L$, M is Meteor and BS stands for BertScore. Higher is better.}
\label{tab:gpt_results_short}
\scalebox{1}{
\begin{tabular}{@{}lllll@{}}
\toprule
                       & \multicolumn{2}{c}{\textbf{Input}}                                & \multicolumn{2}{c}{\textbf{Output}}                              \\ \cmidrule(l){2-5} 
                       & \multicolumn{1}{c}{\textbf{EMA}} & \multicolumn{1}{c}{\textbf{R}} & \multicolumn{1}{c}{\textbf{R}} & \multicolumn{1}{c}{\textbf{BS}} \\ \midrule
\textbf{GPT3.5 + demo} & 22.3                             & 36.9                           & 32.5                           & 87.2                            \\
\textbf{GPT3.5 + CoT}   & {24.8}                    & {45.0}                  & {42.3}                  &  {88.4}     \\           
\textbf{GPT3.5 + FT}   & \textbf{32.6}                    & \textbf{55.9}                  & \textbf{53.6}                  & \textbf{90.6}                   \\ \midrule
\textbf{GPT-4 + CoT}   & 26.7                             & 51.4                           & 52.3                           & 88.9                            \\ \midrule \midrule
\textbf{Human}         & \textit{48.6}                    & \textit{85.5}                  & \textit{58.2}                  & \textit{97.1}                   \\ \bottomrule
\end{tabular}
}

\end{table}

\paragraph{T5 and Flan-T5}

The performance analysis of T5-based methods, summarized in Table \ref{tab:t5_results}, reveals subpar outcomes, highlighting the task's complexity. The random splits setting slightly outperforms the distribution-based setting. T5's generalization issues lead to a 21\% drop in exact match accuracy and a 17\% decrease in Bleu scores. Neither pretrained T5 nor Flan T5 managed meaningful predictions without fine-tuning. Flan-T5, however, generally scores better than T5, with improved BertScores in both settings, indicating enhanced fluency, semantic coherence, and contextual relevance of the outputs.

\paragraph{GPT models}

GPT-3.5 in 1-shot learning achieves a 22.3\% an EMA score on our test set, while T5 and Flan T5 score 12.9\% and 13.8\%, respectively. The input Rouge$_L$ score is 36.9. A closer inspection reveals that while inputs and outputs are semantically correct, the models often fail to follow the set rules, such as wrongly including tools or locations, or misinterpreting instructions like \textit{"dissolve yeast in water"}. However, the fine-tuned GPT-3.5 corrects some of these issues, achieving a 32.6\% exact match score and the highest BertScore at 90.6.

GPT-4+CoT processing has a 26.7\% EMA score, with improved input and output Rouge$_L$ scores of 51.4 and 52.3, respectively, but does not outperform the fine-tuned GPT-3.5. Its BertScore of 88.9 indicates well-contextualized outputs but falls short of the benchmark set by GPT-3.5+FT.

\paragraph{Human evaluation}
\Cref{tab:human_eval} summarizes the models performance according to three different criteria, difficult to measure with automatic metrics. Specifically, the aim is to measure how complete, accurate, and consistent the generated pairs are, and how it compared to references. Flan T5+FT exhibits lower scores across all metrics, particularly in validity. GPT-3.5 demonstrates best performance, achieving perfect consistency and high scores in both completeness and validity. Consequently, it achieves a higher win rate compared to Flan T5 + FT. However, GPT-4 + CoT performs competitively by showing good scores and the highest win rate, indicating robust performance, though it falls slightly behind GPT-3.5 in Consistency. The human performance, as shown in the last row of the table, significantly outperforms the AI models in terms of completeness, validity, and consistency, with a high win rate and low loss rate.

\paragraph{Human performance}
We asked a third set of crowd-workers to fill out 25 sampled recipe tables and evaluate their performance using the automatic metrics described. Results in \Cref{tab:gpt_results_short} and \Cref{tab:gpt_results} describe human performance with an EMA of 48.6\%, an input Rouge$_L$ of 85.5, and an output Rouge$_L$ of 58.2. With a BertScore of 97.1, human performance constitute the benchmark for this task, emphasizing the gap in achieving human-like comprehension and output in complex tasks.

\textbf{Note:} Although a human EMA of 48.6\% may appear low, it's important to note that EMA is a stringent metric sensitive to minor phrasing variations from a reference, which can substantially impact scores. This sensitivity also accounts for the low output Rouge$_L$ score. Despite some subjectivity in the phrasing of inputs and outputs, the primary goal—assessing the ability to follow instructions requiring commonsense cooking reasoning—remains objective. LLMs often struggle with understanding intermediate steps and adjusting ingredient lists according to cooking instructions, highlighting deficiencies in their commonsense reasoning abilities.

\begin{table*}[ht]
\centering
\caption{Samples of errors in predicting input/output pairs. The correct predictions are in green, acceptable in blue, and the wrong predictions are in red.}
\label{tab:errors}
\scalebox{0.75}{
\begin{tabular}{m{4cm}p{3cm}p{3cm}p{3cm}p{3cm}p{3cm}}
\hline
\multicolumn{1}{c}{\textbf{Instruction}} & \multicolumn{1}{c}{\textbf{Ground-truth}} & \multicolumn{1}{c}{\textbf{T5}} & \multicolumn{1}{c}{\textbf{Flan T5}} & \multicolumn{1}{c}{\textbf{GPT+demo}} & \multicolumn{1}{c}{\textbf{GPT}} \\ \hline
\multicolumn{1}{l|}{} & (tomato sauce; water; oregano; ...) & {\color[HTML]{FE0000} (chicken; oil; cheese)} & {\color[HTML]{FE0000} (all ingredient; oil; cheese)} & {\color[HTML]{3166FF} tomatoes, onion, garlic, oregano,...} & {\color[HTML]{3166FF} (tomato; salt; italian seasoning)} \\ \cline{2-6} 
\multicolumn{1}{l|}{\multirow{-2}{4cm}{\textit{combine all ingredients except for the chicken , oil and cheese in a saucepan.}}} & seasoned tomato sauce & {\color[HTML]{32CB00} sauce} & {\color[HTML]{FE0000} chicken fat} & {\color[HTML]{32CB00} combined tomato sauce mixture} & {\color[HTML]{32CB00} tomato sauce} \\ \hline
\multicolumn{1}{l|}{} & (opt. flour; opt. water; dough) & {\color[HTML]{3166FF} (dough; flour; water)} & {\color[HTML]{FE0000} (mixed ingredients; flour; water)} & {\color[HTML]{3166FF} flour or water} & {\color[HTML]{3166FF} (dough; water or flour)} \\ \cline{2-6} 
\multicolumn{1}{l|}{\multirow{-2}{4cm}{\textit{add more flour or water until you get that consistency}}} & dough & {\color[HTML]{32CB00} flour added to dough} & {\color[HTML]{32CB00} flour or water added to mixed ingredients} & {\color[HTML]{FE0000} adjusted consistency} & {\color[HTML]{32CB00} water or flour added to dough} \\ \hline
\multicolumn{1}{l|}{} & (...; salt; blue cornmeal) & {\color[HTML]{FE0000} (dough; olive oil)} & {\color[HTML]{FE0000} (peanut butter; fresh dill; ...)} & {\color[HTML]{FE0000} ingredients} & {\color[HTML]{FE0000} ingredients} \\ \cline{2-6} 
\multicolumn{1}{l|}{\multirow{-2}{4cm}{\textit{place everything in the bowl of an electric mixer with a dough hook}}} & partially mixed blue dough & {\color[HTML]{32CB00} dough} & {\color[HTML]{FE0000} peanut butter mixture} & {\color[HTML]{FE0000} ingredients in the bowl} & {\color[HTML]{FE0000} ingredients} \\ \hline
\multicolumn{1}{l|}{} & NA & {\color[HTML]{FE0000} (chicken; fork)} & {\color[HTML]{32CB00} NA} & {\color[HTML]{32CB00} NA} & {\color[HTML]{FE0000} (fork; chicken)} \\ \cline{2-6} 
\multicolumn{1}{l|}{\multirow{-2}{4cm}{\textit{before putting the chicken on , get a fork}}} & NA & {\color[HTML]{FE0000} chicken with fork} & {\color[HTML]{32CB00} NA} & {\color[HTML]{32CB00} NA} & {\color[HTML]{FE0000} chicken and fork} \\ \hline
\end{tabular}}
\end{table*}

\section{Analysis \& Discussion}
The evaluation of baseline performance highlighted areas needing improvement. We analyze the results to identify the model's strengths and weaknesses, gaining insights into its limitations. We categorize the common errors into three classes: (1) missing or incorrect predictions, (2) inclusion of non-comestibles, and (3) non-descriptive predictions.

\paragraph{Missing descriptions} We noticed that this type of error occurs primarily in the T5 baseline. More specifically, the model will fail in predicting the masked tokens for longer recipes. 
This is a type of error that is frequent in the predictions in all settings and baselines. Some examples of what qualifies as wrong description are shown in \Cref{tab:errors}. 
Given the instruction \textit{combine all ingredients except for the chicken, oil and cheese in a saucepan} with ground truth input \textit{(tomato sauce; water; oregano; basil; thyme; garlic powder; salt; black pepper; bay leaf; lemon juice)} and output \textit{seasoned tomato sauce}, T5 predicts as input \textbf{(chicken; oil; cheese)} and output \textbf{sauce}.
This example demonstrates a fundamental misunderstanding, failing to recognize that explicitly mentioned ingredients should be excluded and incorrectly linking the output to general terms from the instructions rather than deriving it logically from the input.

\paragraph{Presence of non-comestibles}
 The last row of \Cref{tab:errors} clearly illustrates the problem of non-comestibles in generations, where both GPT-3.5 and T5+FT incorrectly include items like \textbf{(chicken; fork)} in the input, leading to an illogical \textbf{chicken with fork} as the output. This issue, labeled as "presence of non-comestibles," appears in 27\% and 15\% of cases for GPT-3.5 and GPT-4 respectively, often occurring in the input column—e.g., the input "\textbf{oven}" for the action "\textbf{preheat}" in the instruction "\textit{preheat the oven at x degrees.}" Providing additional demonstrations reduces the frequency of this error to 12\% for GPT-3.5 and 7\% for GPT-4.

\paragraph{Non-descriptive predictions} GPT-3.5 typically predicts generic terms like \textbf{ingredients} and \textbf{ingredients in the bowl} as input and output for the instruction \textit{place everything in the bowl of an electric mixer with a dough hook}, which, while not incorrect, lacks detail about the nature of the intermediate food. Similarly, experiments with GPT-4 + CoT demonstrate that although the model can generate plausible descriptions, it often avoids genuine reasoning about cooking actions and ingredients, preferring to 'guess' likely phrases. Recipes with precise details expose the limitations of current models, highlighting the need for specialized commonsense reasoning research.

It is worth noting that BertScore, consistently higher than other metrics, suggests that the predicted pairs are semantically very similar to the references, representing an upper evaluation bound. At the same time, EMA, which strictly measures the match between two strings, is a lower bound.

\section{Related Work}
\paragraph{Food understanding} Large datasets in the cooking domain, such as Food-101 \cite{bossard_food-101mining_2014} and Recipe1M \cite{salvador_learning_2017}, have recently lead to significant advancements in food understanding. While these datasets are commonly used as benchmarks in computer vision (e.g., see \cite{pham2021chef}), text-based natural language processing studies have explored areas like recipe text understanding via flow-graphs \cite{mori_flowgraph2text_2014,mori_flow_2014} and recipe parsing \cite{chang_recipescape_2018,jermsurawong_predicting_2015,kiddon_mise_2015}. We introduce a dataset that provides detailed input-output pairs and state changes in the cooking process in natural language, through step-by-step annotations. This detailed information is designed to develop models that can reason and articulate their reasoning paths at fine-grained level, an important feature for generating safe and coherent recipes. 



\paragraph{Decomposing multi-step reasoning tasks}
 allows the model to focus on specific aspects of the task and to gradually build an understanding of the overall problem. 
One such prompting approach is the chain of thought prompting \cite{wei2022chain}, which prompts the language
model to generate a series of intermediate steps that improve the reasoning capabilities in LLMs. \citeauthor{wang2022self} took another step forward and sampled multiple reasoning paths and selected the most relevant output using majority voting. \citeauthor{kojima2022large} further improved the reasoning of LLMs in a zero-shot manner by appending \textit{“Let’s think step by step”} to the prompt. 
In contrast, our work explicitly asks the model to reason by trying to solve the sub-question at a fine-grained level. Most similar to our work is the work of \citeauthor{zhou2022least} which decomposes questions into sub-questions and asks the language model to solve each sub-question sequentially.

\paragraph{Tabular data in LLMs} The reasoning process involves decomposing a multi-step reasoning task is inherently structured. While conventional natural language texts are generated in a 1-dimensional sequential order, the table has a 2-dimensional structure, which allows to reason horizontally and vertically at the same time. We argue that these features justify the choice of tabular data for the proposed task. 
To use an LLM for tabular data, the table must be serialized into a natural text representation. Proposed serialization formats include the simple list or sentence serializations \cite{narayan2022can,borisov2022deep}. \citet{yin2020tabert} also included the column data type in the serialized string. We used the serialization method by \citet{wu-etal-2022-text-table}.

\section{Conclusion}
In this work, we introduce PizzaCommonSense, a dataset for evaluating models' understanding of cooking instructions, focusing on implicit ingredient transformations. We set baselines and demonstrate the challenging nature of the task through evaluations of LLMs. Our experiments underscore the significant limitations of LLMs when applied to this dataset, highlighting their inability to effectively handle complex reasoning that necessitates commonsense knowledge. These findings emphasize the need for advancements in model architecture and training methodologies to address these challenges. Overcoming these hurdles will enable the development of improved models capable of better understanding commonsense reasoning with procedural texts, with potential applications including autonomous agents using procedural information in scientific articles, industrial processes, DIY, as well as cooking. 

\section*{Limitations}
(1) One of the limitations of our dataset is that we collect only one interpretation for each instruction. While this is currently the case, we aim to actively explore strategies to expand the dataset with multiple interpretations, allowing for a richer and more nuanced understanding of the diverse ways to interpret instructions. 

(2) The current scope of our dataset limits its comprehensiveness, as it exclusively encompasses recipes with "pizza" in the title. We leave the expansion to other types of recipe in a future work. (3) While BLEU, ROUGE, and METEOR are widely used metrics for evaluating text, they have certain limitations.
One limitation is that they focus on n-gram overlap, which means they only consider how many words or phrases match between the generated text and the reference text. Finally, they are not able to capture the overall meaning or gist of the text. This means that a model can generate text that is factually accurate but does not convey the same meaning as the reference text. (4) Finally, there are additional concerns that need to be considered which are the bias towards certain cuisines. Most of the recipes are based on western cuisine, specifically from the USA. The recipes show a substantial use of proprietary ingredients which might be a limitation for generalization abilities.

\section*{Ethical considerations}

\paragraph{Data Collection}
We performed the data collection using Amazon Mechanical Turk and data evaluation on CloudConnect Research. We made
sure annotators were fairly compensated by calculating an average hourly wage above the US minimum wage. We maintain the anonymity of all data, ensuring that no personally identifiable information is captured from crowd workers. 
We rigorously curate the tasks and prompts used for data collection, meticulously avoiding any controversial or sensitive topics. This approach minimizes the potential for harm or misuse of the dataset.

\paragraph{Generative models} The generative models are based on pre-trained language models, which may generate offensive content if prompted with inappropriate inputs.

\section*{Acknowledgements}

This research was supported by the Leverhulme Trust grant for the project 'Repurposing of Resources: from Everyday Problem Solving through to Crisis Management' (RPG-2021-182). We also thank all reviewers for their insightful feedback.

\bibliography{anthology,custom,zotero}
\bibliographystyle{acl_natbib}

\appendix

\section{Dataset and Evaluation}
\begin{itemize}
\item Recipe1M : \url{http://im2recipe.csail.mit.edu/dataset/download}
\item ROUGE, BLEU, Meteor: \url{https://github.com/salaniz/pycocoevalcap}
\item BERTScore: \url{https://github.com/huggingface/evaluate}
\end{itemize}

\section{Models and data source} 
T5 and Flan-T5 are available on HuggingFace \footnote{https://huggingface.co/docs/transformers/}. GPT-3.5\footnote{https://platform.openai.com/docs/models/gpt-3-5} and GPT-4\footnote{https://platform.openai.com/docs/models/gpt-4-turbo-and-gpt-4} were accessed from the OpenAi API. our dataset was built from Recipe1M \footnote{http://pic2recipe.csail.mit.edu/} which is publicly available. 

\section{Additional tables}
\Cref{tab:key_statistics} represents the summary of PizzaCommonsense statistics. 
\begin{table}[h]
\centering
\scalebox{0.9}{
\begin{tabular}{@{}lc @{}}
\toprule
\multicolumn{1}{c}{\textbf{Property}} & \multicolumn{1}{c}{\textbf{Value}} \\ \midrule
\# recipes                             & 1087   \\
\# instances & 13141 \\
\# words per instruction (average/median)   & 7.45 / 6.0                         \\
\# words per input (average/median)         & 1.6 / 1.0                          \\
\# words per output (average/median)        & 1.1 / 1.0                          \\
\# instructions per recipe (average/median)        & 12.5 / 11.0                        \\ \bottomrule
\end{tabular}}
\caption{Core Statistics of PizzaCommonSense.}
\label{tab:key_statistics}
\end{table}

\Cref{tab:gpt_results} shows the (complete) quantitative evaluation for GPT-3.5+demo, GPT-3.5+FT and GPT-4+CoT. 

\begin{table*}[h]
\centering
\caption{Quantitative evaluation for GPT-3.5+demo, GPT-3.5+FT and GPT-4+CoT. EMA is exact matching accuracy, B is Bleu, R is Rouge$_L$, M is Meteor and BS stands for BertScore. Higher is better.}
\label{tab:gpt_results}
\scalebox{0.9}{
\begin{tabular}{@{}ccc@{\hskip 30pt}cccc@{}}
\toprule
                       & \multicolumn{2}{c}{\textbf{Input}}                                & \multicolumn{4}{c}{\textbf{Output}}                                                                                                \\ \cmidrule(l){2-7} 
                       & \multicolumn{1}{c}{\textbf{EMA}} & \multicolumn{1}{c}{\textbf{R}} & \multicolumn{1}{c}{\textbf{R}} & \multicolumn{1}{c}{\textbf{B}} & \multicolumn{1}{c}{\textbf{M}} & \multicolumn{1}{c}{\textbf{BS}} \\ \midrule
\textbf{GPT3.5 + demo} & 22.3                             & 36.9                           & 32.5                           & 9.20                           & 20.0                           & 87.2                            \\
\textbf{GPT3.5 + FT}   & \textbf{32.6}                    & \textbf{55.9}                  & \textbf{53.6}                  & \textbf{15.8}                  & \textbf{46.2}                  & \textbf{90.6}                   \\ \midrule
\textbf{GPT-4 + CoT}   & 26.7                             & 51.4                           & 50.9                           & 9.14                           & 40.7                           & 88.9                            \\ \midrule \midrule
\textbf{Human}         & \textit{48.6}                    & \textit{85.5}                  & \textit{58.2}                  & \textit{86.3}                  & \textit{77.1}                  & \textit{97.1}                   \\ \bottomrule
\end{tabular}
}

\end{table*}

\section{Overlap Quantification}

There is some overlap between individual cooking instructions in 3 sets. This is inevitable given that they are all pizza recipes which typically involves a limited set of steps: creating the base, applying sauce, adding toppings, and baking/chilling/setting. This is added to the fact that the task consist of describing “what we had before the action” and “what we get after the action”. Under this setting, our goal is to investigate if models can correctly reason even if the action is the same but the initial set of ingredients. We argue that it is the similar rationale of arithmetic reasoning datasets. The objective is to study compositional aspects of recipe instructions. Each stage (base, sauce, toppings) acts as a distinct "building block," allowing for easier segmentation and analysis of the construction and reasoning process.

The percentage overlap between training and validation sets for instructions is 5.47\%, and 9.21\% between training and testing. This overlap translates to ingredient lists with 3.91\% and 6.26\% overlap for training-validation and training-test sets, respectively. Similarly, the overlap in final outputs (resulting dishes) is 4.12\% and 6.13\% for training-validation and training-test sets. These duplicate instructions are mainly of the type "preheat the over at x" or "serve immediately" (some of the most common). In most cases, the instruction of the first type have as input "NA".

\section{Example of recipe}
\textbf{Title:} White Pizza Triscuit Crackers \\
\textbf{Ingredients:} 1/2 cup frozen chopped broccoli, thawed, drained. (or can even try with chopped thawed spinach), 1/2 cup part-skim ricotta cheese, 1/2 cup shredded mozzarella cheese (or other blend), 1/4 cup parmesan cheese or 14 cup romano cheese, grated topping, 1/2 teaspoon dried oregano, 1/2 teaspoon garlic powder, 48 Triscuit crackers. \\
\textbf{Instructions:} Preheat oven to 350 degrees. Mix broccoli, cheeses and seasonings. Spread 1 teaspoons of the cheese mixture onto each of the 48 crackers. Place on baking sheet. Bake 5 minutes or until hot and bubbly., Serve warm.

\textbf{Masked serialized table:} \\
{\small
\texttt{instructions <s> input <s> actions <s> output <n>} \\ \texttt{preheat oven to 350 degrees . <s> \textcolor{blue}{<in 0>} <s> preheat <s> \textcolor{red}{<out 0>}} \\ \texttt{mix broccoli , cheeses and seasonings . <s> \textcolor{blue}{<in 1>} <s> mix <s> \textcolor{red}{<out 1>}} \\ \texttt{spread 1 teaspoons of the cheese mixture onto each of the 48 crackers . <s> \textcolor{blue}{<in 2>} <s> spread <s> \textcolor{red}{<out 2>}} \\ \texttt{place on baking sheet . <s> \textcolor{blue}{<in 3>} <s> place <s> \textcolor{red}{<out 3>}} \\ \texttt{bake 5 minutes or until hot and bubbly . <s> \textcolor{blue}{<in 4>} <s> bake <s> \textcolor{red}{<out 4>}} \\ \texttt{serve warm . <s> \textcolor{blue}{<in 5>} <s> serve <s> \textcolor{red}{<out 5>}}}

\textbf{Ground truth serialized table:} \\
{\small \texttt{instructions <s> input <s> actions <s> output <n>} \\ \texttt{preheat oven to 350 degrees . <s> \textcolor{blue}{NA} <s> preheat <s> \textcolor{red}{NA}} \\ \texttt{mix broccoli , cheeses and seasonings . <s> \textcolor{blue}{(broccoli; ricotta cheese; mozzarella cheese; parmesan cheese; seasonings)} <s> mix <s> \textcolor{red}{broccoli and cheeses mixture}} \\ \texttt{spread 1 teaspoons of the cheese mixture onto each of the 48 crackers . <s> \textcolor{blue}{(broccoli and cheeses mixture; crackers)} <s> spread <s> \textcolor{red}{broccoli and cheeses mixture spread over crackers}} \\ \texttt{place on baking sheet . <s> \textcolor{blue}{broccoli and cheeses mixture spread over crackers} <s> place <s> \textcolor{red}{topped crackers on baking sheet}} \\ \texttt{bake 5 minutes or until hot and bubbly . <s> \textcolor{blue}{topped crackers on baking sheet} <s> bake <s> \textcolor{red}{baked crackers}} \\ \texttt{serve warm . <s> \textcolor{blue}{baked crackers} <s> serve <s> \textcolor{red}{served crackers}}  \\  
}
\section{Example of predictions by T5}
\label{sec:t5_prediction}
{\small
\texttt{instructions <s> input <s> actions <s> output <n>} \\ \texttt{preheat oven to 350 degrees . <s> \textcolor{blue}{na} <s> preheat <s> \textcolor{red}{na}} \\ \texttt{mix broccoli , cheeses and seasonings . <s> \textcolor{blue}{(broccoli; cheese; seasonings)} <s> mix <s> \textcolor{red}{broccoli cheese mixture}} \\ \texttt{spread 1 teaspoons of the cheese mixture onto each of the 48 crackers . <s> \textcolor{blue}{(broccoli cheese mixture; crackers)} <s> spread <s> \textcolor{red}{crackers with broccoli cheese mixture}} \\ \texttt{place on baking sheet . <s> \textcolor{blue}{crackers with broccoli cheese mixture} <s> place <s> \textcolor{red}{crackers with broccoli cheese mixture}} \\ \texttt{bake 5 minutes or until hot and bubbly . <s> \textcolor{blue}{crackers with broccoli cheese mixture} <s> bake <s> \textcolor{red}{baked crackers with broccoli cheese mixture}} \\ \texttt{serve warm . <s> \textcolor{blue}{baked crackers with broccoli cheese mixture} <s> serve <s> \textcolor{red}{served broccoli cheese crackers}} 
}

\section{Example of predictions by Flan-T5}
\label{sec:flan_prediction}
{\small
\texttt{instructions <s> input <s> actions <s> output <n>} \\ \texttt{preheat oven to 350 degrees . <s> \textcolor{blue}{na} <s> preheat <s> \textcolor{red}{na}} \\ \texttt{mix broccoli , cheeses and seasonings . <s> \textcolor{blue}{(broccoli; cheeses; seasonings)} <s> mix <s> \textcolor{red}{cheese mixture}} \\ \texttt{spread 1 teaspoons of the cheese mixture onto each of the 48 crackers . <s> \textcolor{blue}{(cheese mixture; crackers)} <s> spread <s> \textcolor{red}{crackers with cheese mixture}} \\ \texttt{place on baking sheet . <s> \textcolor{blue}{crackers with cheese mixture} <s> place <s> \textcolor{red}{crackers with cheese mixture}} \\ \texttt{bake 5 minutes or until hot and bubbly . <s> \textcolor{blue}{crackers with cheese mixture} <s> bake <s> \textcolor{red}{baked crackers}} \\ \texttt{serve warm . <s> \textcolor{blue}{baked crackers} <s> serve <s> \textcolor{red}{served warm crackers}} 
}

\section{Example of predictions by GPT-3.5}
\label{sec:gpt_van_prediction}
{\small
\texttt{It seems that we are preparing a broccoli and cheese mixture to spread on crackers and then bake. Here are the completions: \\ <in 0>: NA \\- <out 0>: NA\\- <in 1>: (broccoli, ricotta cheese, mozzarella cheese, parmesan cheese, seasonings)\\- <out 1>: broccoli and cheeses mixture\\- <in 2>: (broccoli and cheeses mixture, crackers)\\- <out 2>: broccoli and cheeses mixture spread over crackers\\- <in 3>: broccoli and cheeses mixture spread over crackers\\- <out 3>: topped crackers on baking sheet\\- <in 4>: topped crackers on baking sheet\\- <out 4>: baked crackers\\- <in 5>: baked crackers\\- <out 5>: served crackers}

}

\section{Example of predictions by GPT-3.5 with demonstrations}
\label{sec:gpt_demo_prediction}
{\small
\texttt{instructions <s> input <s> actions <s> output <n>} \\ \texttt{ preheat oven to 350 degrees   <s> \textcolor{blue}{NA} <s>  preheat  <s> \textcolor{red}{NA}} \\ \texttt{ mix broccoli , cheeses and seasonings   <s> \textcolor{blue}{broccoli, cheeses, seasonings} <s>  mix  <s> \textcolor{red}{mixture of broccoli, cheeses and seasonings}} \\ \texttt{ spread 1 teaspoons of the cheese mixture onto each of the 48 crackers   <s> \textcolor{blue}{mixture of broccoli, cheeses and seasonings} <s>  spread  <s> \textcolor{red}{crackers spread with mixture of broccoli, cheeses and seasonings}} \\ \texttt{ place on baking sheet   <s> \textcolor{blue}{crackers spread with mixture of broccoli, cheeses and seasonings} <s>  place  <s> \textcolor{red}{crackers with mixture of broccoli, cheeses and seasonings on baking sheet}} \\ \texttt{ bake 5 minutes or until hot and bubbly   <s> \textcolor{blue}{crackers with mixture of broccoli, cheeses and seasonings on baking sheet} <s>  bake  <s> \textcolor{red}{hot and bubbly crackers with mixture of broccoli, cheeses and seasonings}} \\ \texttt{ serve warm   <s> \textcolor{blue}{hot and bubbly crackers with mixture of broccoli, cheeses and seasonings} <s>  serve  <s> \textcolor{red}{hot and bubbly crackers with mixture of broccoli, cheeses and seasonings ready to serve}}

}
\section{Example of predictions by GPT-3.5 with fine-tuning}
\label{sec:ft_gpt_prediction}

{\small
\texttt{instructions <s> input <s> actions <s> output <n>} \\ \texttt{ preheat oven to 350 degrees   <s> \textcolor{blue}{NA} <s>  preheat  <s> \textcolor{red}{NA}} \\ \texttt{ mix broccoli , cheeses and seasonings   <s> \textcolor{blue}{(brocolli; cheese; seasoning)} <s>  mix  <s> \textcolor{red}{cheese and broccoli mix}} \\ \texttt{ spread 1 teaspoons of the cheese mixture onto each of the 48 crackers   <s> \textcolor{blue}{(broccoli and cheese mix; cracker)} <s>  spread  <s> \textcolor{red}{cheese and broccoli mixture spread}} \\ \texttt{ place on baking sheet   <s> \textcolor{blue}{cheese and broccoli covered cracker} <s>  place  <s> \textcolor{red}{cheese and broccolli crackers on baking sheet}} \\ \texttt{ bake 5 minutes or until hot and bubbly   <s> \textcolor{blue}{cheese and broccolli crackers on baking sheet} <s>  bake  <s> \textcolor{red}{baked cheese and broccolli crackers}} \\ \texttt{ serve warm   <s> \textcolor{blue}{baked cheese and broccolli crackers} <s>  serve  <s> \textcolor{red}{broccoli and cheese crackers}}
}

\section{Example of predictions by GPT-4}
\label{sec:gpt_cot_prediction}

{\small
\texttt{<preheat oven to 350 degrees> <s> <in0> <s> preheat <s> NA }\\\texttt{<mix broccoli, cheeses and seasonings> <s> <in1> <s> mix <s> cheese mixture } \\ \texttt{<spread 1 teaspoons of the cheese mixture onto each of the 48 crackers> <s> <in2> <s> spread <s> cheese-topped crackers }\\ \texttt{<place on baking sheet> <s> <in3> <s> place <s> crackers on baking sheet} \\ \texttt{<bake 5 minutes or until hot and bubbly> <s> <in4> <s> bake <s> baked cheese-topped crackers }\\ \texttt{<serve warm> <s> <in5> <s> serve <s> served cheese-topped crackers <n>}}

\section{Example of predictions by GPT-4+CoT}
\label{sec:gpt_van_prediction}

{\small
\texttt{instructions <s> input <s> action <s> output <n>} \\ \texttt{preheat oven to 350 degrees  <s> oven <s> preheat <s> preheated oven} \\ \texttt{mix broccoli , cheeses and seasonings  <s> broccoli, cheeses, seasonings <s> mix <s> broccoli cheese mix } \\ \texttt{ spread 1 teaspoons of the cheese mixture onto each of the 48 crackers  <s> cheese mixture, crackers <s> spread <s> cheese-topped crackers} \\ \texttt{ place on baking sheet  <s> cheese-topped crackers <s> place <s> cheese-topped crackers on baking sheet} \texttt{ bake 5 minutes or until hot and bubbly  <s> cheese-topped crackers on baking sheet <s> bake <s> baked cheese-topped crackers } \\ \texttt{serve warm  <s> baked cheese-topped crackers <s> serve <s> served baked cheese-topped crackers }
}
\begin{figure*}[!t]
    \centering
    \includegraphics[angle=90, width=0.87\textwidth]{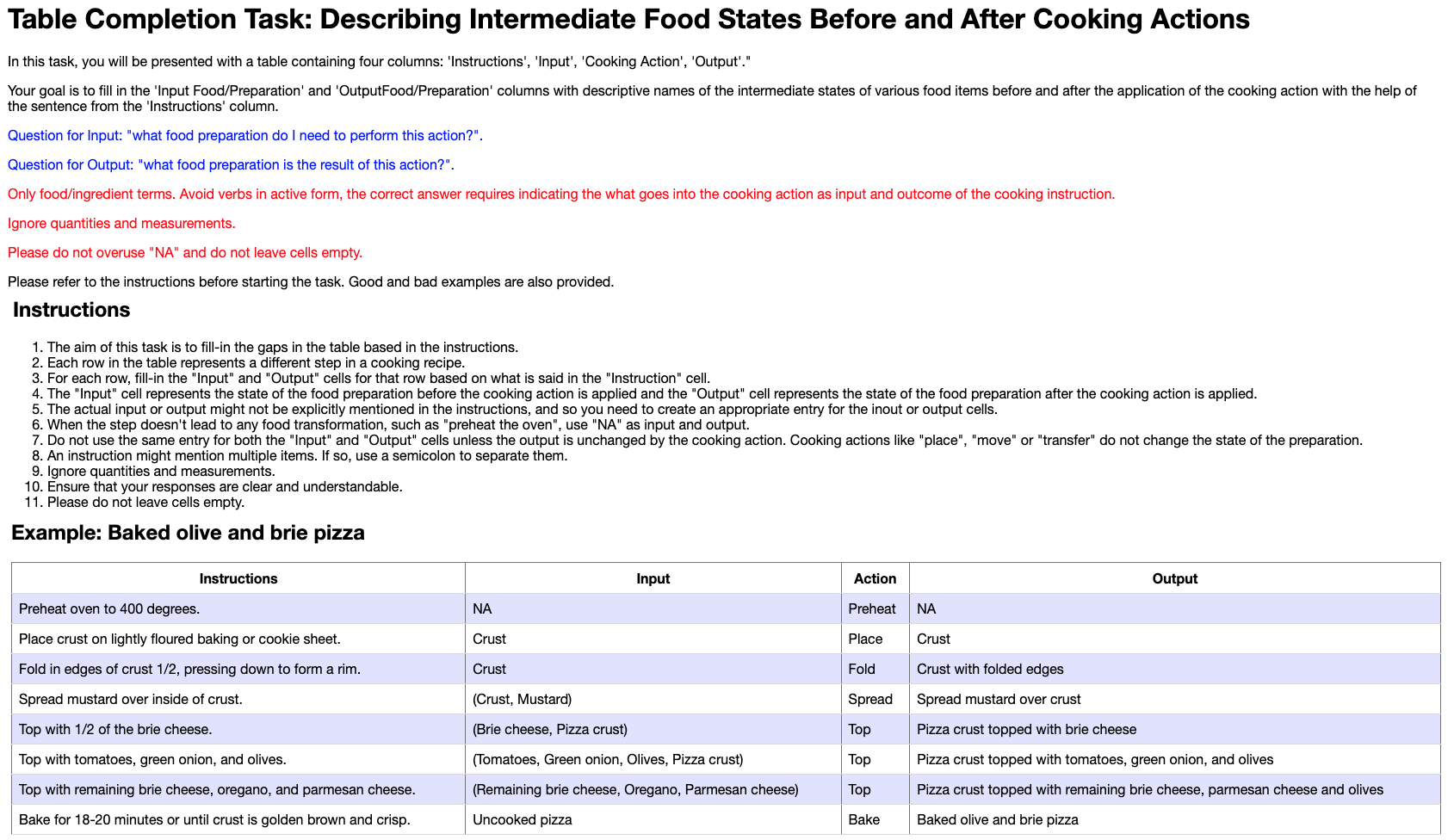}
    \caption{Data collection interface on AMT.}
    \label{fig:amt_collection}
\end{figure*}

\end{document}